\documentclass[reprint,amsmath,amssymb,aps,prl]{revtex4-1}
\usepackage[utf8]{inputenc}
\usepackage{graphicx}
\usepackage{dcolumn}
\usepackage{bm}
\usepackage[space]{grffile}
\usepackage[export]{adjustbox} 
\usepackage{booktabs}
\usepackage[caption=false]{subfig}
\usepackage{natbib}
\usepackage{lineno}

\begin{document}

\title[The Distributed Information Bottleneck]{The Distributed Information Bottleneck reveals the explanatory structure of complex systems}

\author{Kieran A. Murphy$^{1}$}

\author{Dani S. Bassett$^{1,2,3,4,5,6,7}$}

\affiliation{
    $^{1}$Department of Bioengineering, School of Engineering and Applied Science, University of Pennsylvania, Philadelphia, PA 19104, USA
}
\affiliation{
	$^{2}$Department of Electrical \& Systems Engineering, School of Engineering \& Applied Science, University of Pennsylvania, Philadelphia, PA 19104, USA}
\affiliation{
	$^{3}$Department of Neurology, Perelman School of Medicine, University of Pennsylvania, Philadelphia, PA 19104, USA
}
\affiliation{
	$^{4}$Department of Psychiatry, Perelman School of Medicine, University of Pennsylvania, Philadelphia, PA 19104, USA
}
\affiliation{
	$^{5}$Department of Physics \& Astronomy, College of Arts \& Sciences, University of Pennsylvania, Philadelphia, PA 19104, USA
}
\affiliation{
	$^{6}$The Santa Fe Institute, Santa Fe, NM 87501, USA
}
\affiliation{
$^{7}$To whom correspondence should be addressed: dsb@seas.upenn.edu
}

\maketitle
\textbf{
    The fruits of science are relationships made comprehensible, often by way of approximation.
    While deep learning is an extremely powerful way to find relationships in data,
    its use in science has been hindered by the difficulty of understanding the learned relationships.
    The Information Bottleneck (IB) \cite{tishbyIB2000} is an information theoretic framework for understanding a relationship between an input and an output in terms of a trade-off between the fidelity and complexity of approximations to the relationship.
    Here we show that a crucial modification---distributing bottlenecks across multiple components of the input---opens fundamentally new avenues for interpretable deep learning in science.
    The Distributed Information Bottleneck throttles the downstream complexity of interactions between the components of the input, deconstructing a relationship into meaningful approximations found through deep learning without requiring custom-made datasets or neural network architectures.
    Applied to a complex system, the approximations illuminate aspects of the system's nature by restricting---and monitoring---the information about different components incorporated into the approximation while maximizing predictability of a related quantity.
    We demonstrate the Distributed IB's explanatory utility in systems drawn from applied mathematics and condensed matter physics. 
    In the former, we deconstruct a Boolean circuit into approximations that isolate the most informative subsets of input components without requiring exhaustive search. 
    In the latter, we localize information about future plastic rearrangement in the static structure of a sheared glass, and find the information to be more or less diffuse depending on the system's preparation.
    By way of a principled scheme of approximations, the Distributed IB brings much-needed interpretability to deep learning and enables unprecedented analysis of information flow through a system.
}

\section{Introduction}
Science is built upon the understanding of relationships: e.g., the evolution of the future from the present, the connection between form and function, and the downstream effects of an intervention are broad classes of relationships with a thematic ubiquity in science.
For all of deep learning's remarkable ability to find complex relationships in data, its use in science has been impeded due to significant issues of interpretability \cite{rudin2022interpretable,fan2021interpretabilityreview,molnar2022interpretableML}.
When understanding is the ultimate goal, rather than performance, interpretability is absolutely essential.

The key to interpretability is successive approximation, allowing detail of a relationship to be incorporated gradually so that humans' limited cognitive capacity \cite{cowan2010workingmemory,rudin2022interpretable} does not prevent comprehension of the full relationship.
Interpretability in machine learning may be achieved through architectural constraints that simplify the function space to search (e.g., generalized additive models \cite{gam2021,lengerich2020purifying,agarwal2021NAM}, decision trees \cite{quinlan1986ID,breiman2017CART,quinlan2014c45}, and support vector machines \cite{cortes1995svm,schoenholz2016natphys}), so that comprehensible components combine in a straightforward manner.
When deep neural networks are involved, however, interpretability generally takes the reduced form of post-hoc explainability \cite{rudin2022interpretable,samek2021explainabilityrev,ribeiro2016LIME}, for example through limited-scope feature ablation or feature attribution methods~\cite{sundararajan2017axiomatic,olah2018distill,raghu2020survey}.
We seek a solution that maintains interpretability of the learned relationships while leveraging the full complexity of deep neural networks.

To the extent that a relationship found in data with machine learning mirrors the underlying relationship in the natural world, interpretability becomes insight about the system under study.
An approximation scheme becomes a series of ``broad strokes'' recreations of the relationship and a prioritization of detail.
Complex systems serve as rich objects of study because diverse behavior at the largest scales arises with sensitive dependence on detail at the smallest scales, due to repeated interactions between a multitude of simpler components~\cite{anderson1972more}.
These systems are natural targets for interpretable deep learning, where approximating micro to macro relationships illuminates the most relevant details of the system and the nature of the system's complexity.

The Information Bottleneck (IB) is a promising framework to lend interpretability to deep learning and allow in-depth analysis of a relationship~\cite{tishbyIB2000,asoodeh2020bottleneck}.
Given random variables $X$ and $Y$ serving as an input and an output, the IB defines a spectrum of compressed representations of $X$ that retain only the most relevant information about $Y$.
The potential of IB to analyze relationships was recently strengthened through connections to the renormalization group in statistical physics, one of the field's most powerful tools~\cite{gordonrelevance2021,kline2021RGIB}.
Although IB serves as a useful framework for examining the process of learning \cite{tishbyDL2015,saxe2019}, it has limited capacity to find useful approximations through optimization, particularly when the relationship between $X$ and $Y$ is deterministic (or nearly so) \cite{kolchinsky2018caveats,kolchinskyNonlinearIB2019}. 
The problem arises from the location of the namesake bottleneck: it occurs after processing the complete input, such that the learned representation may involve arbitrarily complex relationships between the components of the input without penalty.
The result is that much of the spectrum of learned representations is the trivial noisy rendition of a high-fidelity reconstruction~\cite{kolchinsky2018caveats}.

\begin{figure}
    \centering
    \includegraphics[width=0.9\linewidth]{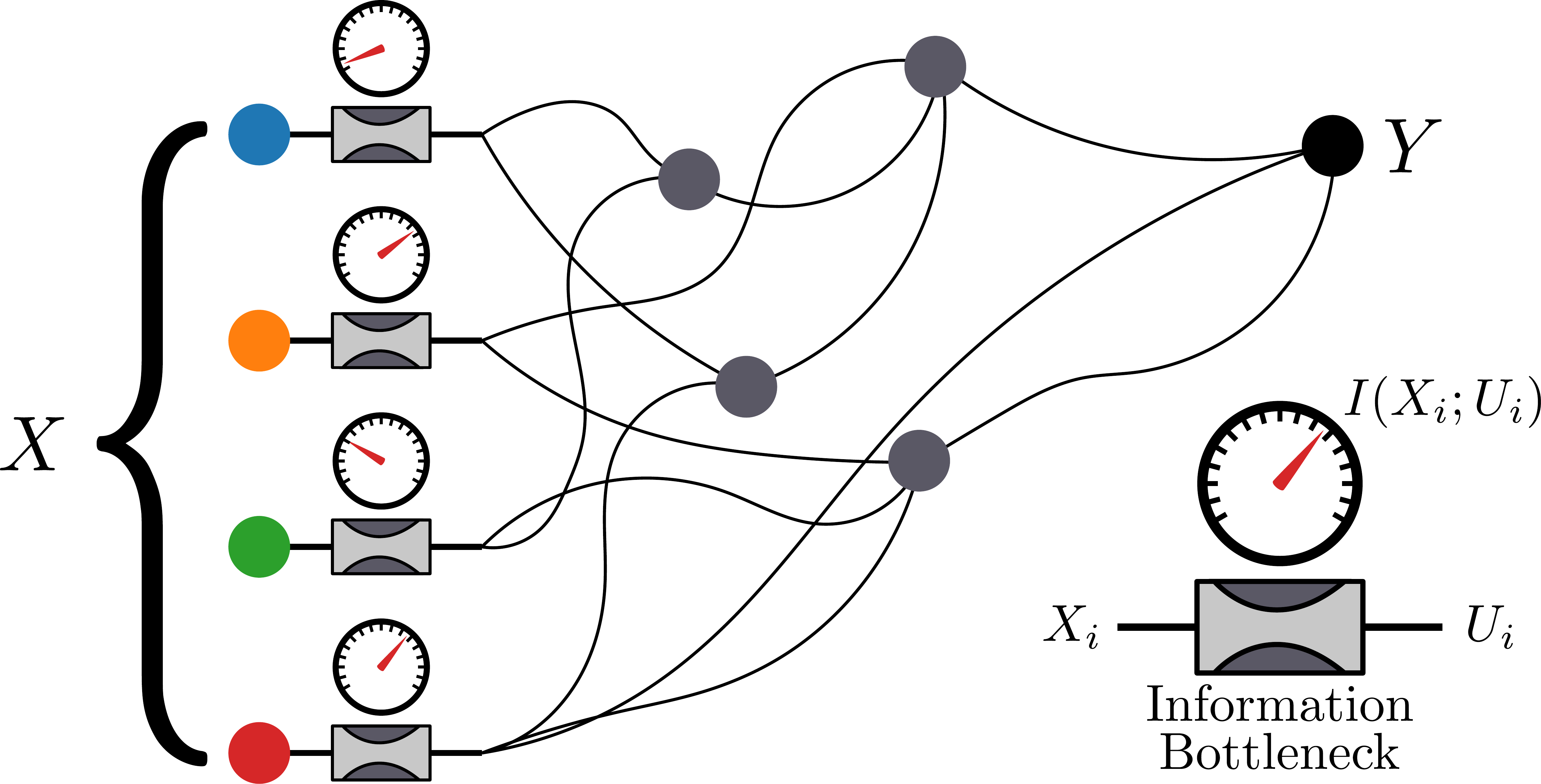}
    \caption{\textbf{Distributed Information Bottleneck for insight into complex relationships.}
    Here, an input $X$ has multiple components $\{X_i\}$ that share some amount of information with an output $Y$.
    The grey nodes represent interaction terms between the components, and the connections between nodes indicate information flow.
    We show in this work that distributing bottlenecks on the information from different components of the input $X$ throttles the downstream complexity of the interactions and yields a continuum of approximations of the relationship between $X$ and $Y$.
    The amount of information passing through each bottleneck---from each $X_i$ into a learned representation $U_i$---reflects the relevance of each component for predicting $Y$, for each level of approximation.
}
    \label{fig:highlevel}
\end{figure}

If bottlenecks are instead distributed after multiple components of the input (Fig.~\ref{fig:highlevel}), information becomes restricted upstream of any interactions between the components (represented by the grey nodes of Fig.~\ref{fig:highlevel}).
Finding the most relevant information in $X$ is then a problem of allocation of information between the components for participation in the most relevant downstream interactions, offering a powerful foothold into the nature of the relationship.
Our central contribution is to show how an IB variant---the Distributed IB~\cite{aguerri2018DIB}, concerning the optimal scheme to integrate information from multiple sources---becomes a powerful diagnostic about a relationship when components of the input serve as distributed sources of information.
A relationship found in data through deep learning is rendered interpretable through a continuous spectrum of approximations, parameterized by the total amount of information allocated across all input components.
The ability to use unconstrained deep learning to meaningfully track information in a relationship opens up fundamentally new analyses of complex systems.

To demonstrate the far-reaching potential of the Distributed IB for interpretable deep learning in science, we apply the framework to three scenarios which represent common motifs in the study of complex systems across disparate fields of science and engineering.
We first study Boolean circuits where the inputs and outputs are binary variables, making the application of the Distributed IB straightforward and allowing relevant information theoretic quantities to be measured directly.
Next we focus on images---relationships between position and color---as they often contain a complex interplay of correlations over multiple length scales and thus serve as a challenging relationship to approximate.
Further, images allow visualization of the entire relationship at once, including the approximate relationships found by the Distributed IB. 
Finally, we gain new insight on an active problem in the physics of amorphous plasticity, analyzing the relationship between static structure and imminent plastic rearrangement in a glass under shear deformation.
We find the most informative markers of the static structure and compare the scheme of approximations found for different quench protocols.
As we navigate these increasingly difficult scenarios, we demonstrate how the Distributed IB illuminates the explanatory structure of complex systems by tracking the flow of information in a relationship.

\section{Methods}
Let $X,Y\sim p(x,y)$ be the random variables constituting the relationship of interest.
The mutual information between two variables is a measure of their statistical dependence, defined as the reduction of entropy between the product of their marginal distributions (as if the variables were independent) and their joint distribution:
\begin{equation} \label{eqn:MI}
    I(X;Y) = H(X) + H(Y) - H(X,Y),
\end{equation}
\noindent with $H(X)=\mathbb{E}_{x\sim p(x)}[-\textnormal{log} \ p(x)]$ Shannon's entropy~\cite{shannon1948mathematical}.

The Information Bottleneck (IB) \cite{tishbyIB2000} probes the relationship between $X$ and $Y$ by way of a rate-distortion problem to convey maximal information from $X$ about $Y$ through a constrained channel, realized as a representation $U=f(X)$.
The representation is found by minimizing a loss consisting of two competing mutual information terms balanced by a scalar parameter $\beta$:
\begin{equation} \label{eqn:IB}
    \mathcal{L}_\textnormal{IB} = \beta I(U;X) - I(U;Y).
\end{equation}
\noindent The first term is the bottleneck, acting as a penalty on information passing into $U$, with $\beta$ determining the strength of the bottleneck.
In the limit where $\beta\rightarrow 0$, the bottleneck is fully open and all information from $X$ may be freely conveyed into $U$ in order for it to be maximally informative about $Y$. 
As $\beta$ increases, only the most relevant information in $X$ about $Y$ becomes worth conveying into $U$, until eventually a trivial, vacuous $U$ is optimal.

Because mutual information is difficult to measure in practice \cite{saxe2019,mcallester2020infolimitations}, the IB objective is tractable only in limited scenarios such as when $X$ and $Y$ are discrete~\cite{tishbyIB2000} or normally distributed~\cite{chechikGaussianIB2005}.
To be practically viable, variants of IB replace the mutual information terms of Eqn.~\ref{eqn:IB} with bounds amenable to deep learning \cite{alemiVIB2016,achillesoatto2018,kolchinskyNonlinearIB2019}.
We follow the Variational Information Bottleneck (VIB) \cite{alemiVIB2016}, which 
learns representations $U$ in a framework nearly identical to that of Variational Autoencoders~\cite{vae,betavae,kingma2019introduction}.
The input $X$ is encoded as a distribution in representation space $p(u|x)=f(x,\phi,\epsilon)$ with a neural network parameterized by weights $\phi$.
A source of noise $\epsilon\sim\mathcal{N}(0,1)$ allows gradient backpropagation in what is commonly referred to as the ``reparameterization trick''~\cite{vae}.
The bottleneck manifests as the Kullback-Leibler (KL) divergence---$D_\textnormal{KL}(w(x)||z(x))=\mathbb{E}_{x\sim w(x)}[-\textnormal{log} \ (z(x)/w(x))]$---between the encoded distribution $p(u|x)$ and a prior distribution $r(u)=\mathcal{N}(0,1)$.
As the KL divergence tends to zero, all representations become indistinguishable from the prior and from each other, and therefore uninformative.
Finally, a representation is sampled from $p(u|x)$ and then decoded to a distribution over the output $Y$ with a second neural network parameterized by weights $\psi$, $q(y|u)=g(u,\psi)$.
The second term of Eqn.~\ref{eqn:IB}, measuring the predictability of $Y$ from the representation $U$, is replaced with the expected cross entropy between the predicted distribution and the ground truth.
In place of Eqn.~\ref{eqn:IB}, the following loss is minimized with standard gradient descent methods:
\begin{equation}\label{eqn:vib}
    \mathcal{L}_\textnormal{VIB} = \beta D_\textnormal{KL}(p(u|x)||r(u)) - \mathbb{E}[\textnormal{log} \ q(y|u)].
\end{equation}
\noindent

\begin{figure*}
    \centering
    \includegraphics[width=\textwidth]{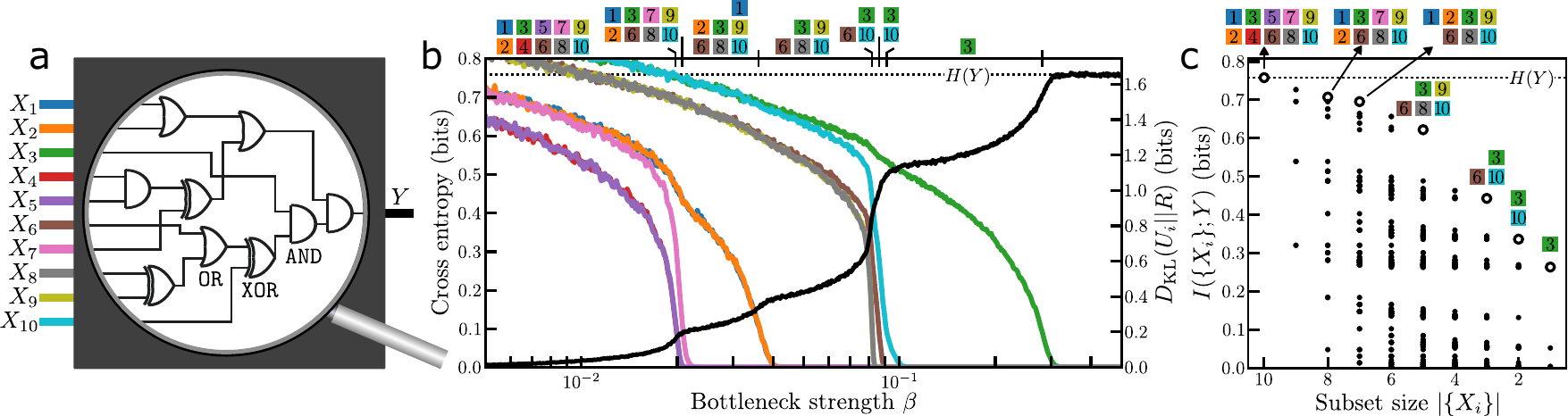}
    \caption{\textbf{Opening a black-box Boolean circuit with the Distributed Information Bottleneck.} \textbf{(a)} A Boolean circuit has ten binary inputs $\{X_i\}$ connected through \texttt{AND}, \texttt{OR}, and \texttt{XOR} gates to one binary output $Y$. 
    \textbf{(b)} With the Distributed IB, each input is compressed and the training objective (Eqn.~\ref{eqn:DIB}) balances predictability of $Y$ with the sum total of information conveyed about each input.
    Sweeping over the bottleneck strength $\beta$ finds a series of relationships between compressed input components and the output $Y$. 
    The cross entropy error of each relationship's prediction of $Y$, shown in black (left vertical axis), is nearly zero when the bottleneck is weakly applied (small $\beta$) and obtains its maximum value, the entropy $H(Y)$ (dotted line), after $\beta \approx 0.3$.
    Information transmitted about each of the inputs (colors corresponding to input gates in panel \textbf{(a)}) is measured through the proxy quantity $D_\textnormal{KL}(U_i||R)$ (right vertical axis).
    Information about the $X_i$ decreases heterogeneously as the bottleneck tightens, with more information allocated to the more relevant components for predicting $Y$.
    Over the course of the $\beta$ sweep, the scheme of approximations of the relationship between $X$ and $Y$ utilizes different subsets of the inputs (those above a threshold $D_\textnormal{KL}(U_i||R)$ are indicated at the top of the plot).
    \textbf{(c)} The mutual information $I(\{X_i\};Y)$ between all subsets of input channels $\{X_i\}$ and the output $Y$ are shown as black circles; there is a large range in the amount of information that different subsets contain with respect to $Y$.
    The maximum mutual information arises from the combination of all ten inputs and the output, equal to the entropy $H(Y)$ (dotted line).
    Every subset of inputs utilized by the Distributed IB in the approximation scheme in \textbf{(b)} is the subset with maximal information for its size (open circles).}
    \label{fig:circuit}
\end{figure*}

The Distributed IB \cite{aguerri2018DIB} has been proposed as a solution to the classic ``CEO problem'', or multiterminal source coding, in information theory ~\cite{berger1996ceo,steiner2021distributedcompression}.
The problem concerns the optimal scheme of compressing multiple sources independently before transmitting to a central decoder to predict some related quantity. 
For example, multiple video cameras independently compress their signals without knowledge of what the other cameras have recorded; the Distributed IB finds the optimal scheme given a constraint on the total transmitted data across all cameras.
Our central contribution is to use the Distributed IB for the analysis of relationships, in which case optimal compression schemes serve as approximations that render the relationship interpretable and illuminate aspects of its nature that are inaccessible to other methods.

Let $\{X_i\}$ be a decomposition of the variable $X$ such that each component is conditionally independent of all others given $X$.
A bottleneck is installed after each $X_i$ by way of a compressed representation $U_i$, and the full set of representations $U_X=\{U_i\}$ is used to predict a variable $Y$.
The scheme is codified in the following loss:
\begin{equation} \label{eqn:DIB}
    \mathcal{L}_\textnormal{DIB} = \beta \sum_i I(U_i;X_i) - I(U_X;Y).
\end{equation}
\noindent The same variational bounds of the Variational IB \cite{alemiVIB2016} can be applied in the Distributed IB setting \cite{aguerriDVIB2021}.

When the output $Y$ is a continuous variable, the cross entropy bound of Eqn.~\ref{eqn:vib} is commonly evaluated by discretizing the support of $Y$ and treating the prediction as a classification problem.
The resolution of the discretization is manually distributed, and the number of outputs must grow rapidly (with the dimension of $Y$) for finer resolution.
We avoid this issue for continuous $Y$ by employing the Noise Contrastive Estimation (InfoNCE) loss used in representation learning as a different bound for the mutual information in Eqn.~\ref{eqn:DIB}, $\mathcal{L}_\textnormal{InfoNCE}\ge- I(U_X;Y)$~\cite{oord2018InfoNCE,poole2019variational}. 
Instead of decoding the combined representation $U_X$ to a distribution over $Y$, we encode $Y$ and compare $U_X$ to $U_Y$ in a shared representation space.
In practice this comparison is evaluated through the following loss contribution:
\begin{equation}\label{eqn:infonce}
    \mathcal{L}_\textnormal{InfoNCE} = -\sum_i^n \textnormal{log} \frac{\textnormal{exp}(s(u_X^{(i)},u_Y^{(i)})/\tau)}{\sum_j^n \textnormal{exp}(s(u_X^{(i)},u_Y^{(j)})/\tau)},
\end{equation}
\noindent where both sums run over a batch of $n$ examples, $s(u,v)$ is a measure of similarity (e.g., negative Euclidean distance), and $\tau$ acts as an effective temperature.  
The form of Eqn.~\ref{eqn:infonce} is equivalent to that of a standard cross entropy loss for identifying the embedding $u_Y^{(i)}$ correspondent to $u_X^{(i)}$ out of all embeddings in a batch $\{u_Y^{(j)}\}$.

In order to obtain a continuum of approximations of the relationship between $X$ and $Y$, a sweep through $\beta$ is made starting with negligible information restrictions so that the relationship between $\{X_i\}$ and $Y$ may be found without obstruction~\cite{wu2020learnability}.
As $\beta$ increases, the evolution of the terms in the loss measures aspects of the approximations.
The KL terms for the different components' representations track the allocation of information across components of $X$ while the cross entropy loss measures the degrading predictive power over $Y$.
Thus we obtain---in addition to the approximate relationships along the continuum---a detailed record of the shifting flow of information from the components of $X$ to $Y$ as the flow gradually decays to vacuity.

\section{Results}
\subsection{Boolean circuit: relation between binary inputs and binary output}
To begin, we consider a Boolean circuit that has ten binary inputs $X=\{X_i\}$, $X_i \in \mathbb{B}$, routing through logical \texttt{AND}, \texttt{OR}, and \texttt{XOR} gates to produce a binary output $Y \in \mathbb{B}$ (Fig.~\ref{fig:circuit}a).
With access only to input-output pairs, we wish to infer properties of the black-box relationship between $\{X_i\}$ and $Y$.
A neural network may readily be trained on the input-output pairs, but this merely creates a new black box to analyze.
We instead seek insight from the outset by finding a scheme of approximations to the relationship.

Accordingly, we consider a Distributed IB optimization for this Boolean circuit as we sweep over $\beta$ (Fig.~\ref{fig:circuit}b).
The error term in the loss---the cross entropy of the prediction with the ground truth---increases in a continuous stepping fashion as the bottleneck strength increases, suggesting multiple robust approximations~\cite{strouse2019robustclusters}.
As $\beta$ increases, the information about the inputs shrinks non-uniformly, as recorded by the ten KL divergence contributions to the loss.
For values of $\beta$ where the error increases quickly, information about a subset of input components is lost.
The sweep over $\beta$ effectively sorts the input gates by their relevance to predicting the output bit $Y$.
The best approximation of the relationship between $\{X_i\}$ and $Y$ includes all ten inputs; the next best discards inputs $X_4$ and $X_{5}$.
Eventually an approximation involving only one input---$X_3$---is found, before all predictability of $Y$ is lost.

\begin{figure*}
    \centering
    \includegraphics[width=\textwidth]{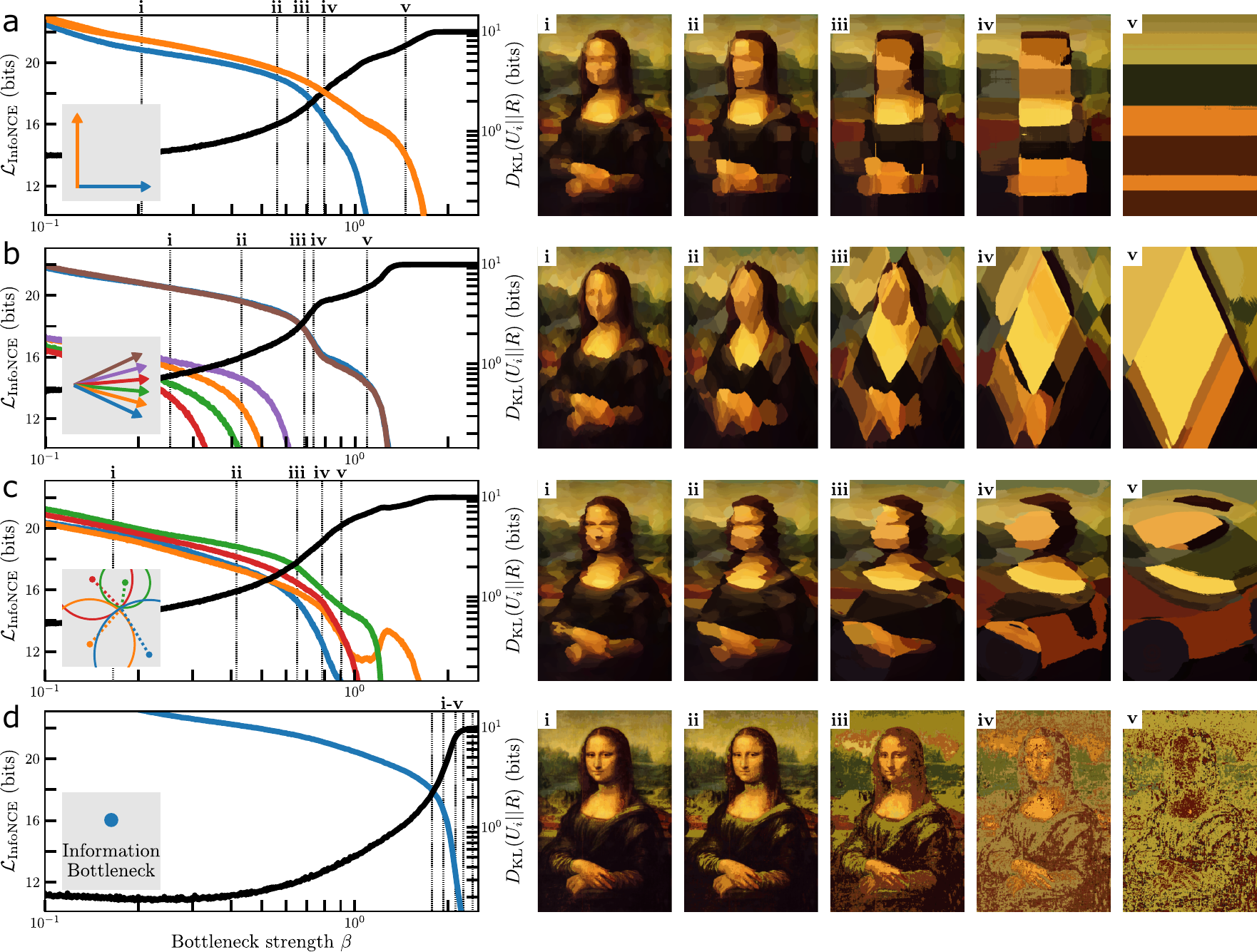}
    \caption{\textbf{Visualizing approximations with different decompositions of the input.} \textbf{(a)} da Vinci's painting of the Mona Lisa represents a specific relationship between the position in the frame ($X$) and color ($Y$). The position is decomposed into horizontal (blue) and vertical (orange) components for use with Distributed IB (inset schematic).  
    We display noteworthy approximations on the right (\textbf{i}-\textbf{v}), whose associated $\beta$ values are marked by the vertical bars in the plot.
    The prediction for each position is a distribution over colors; the color displayed is the one with maximum probability.
    As the bottleneck strength $\beta$ increases, information about each of the components is gradually discarded until the only information comes from the vertical component (approximation \textbf{v}).
    \textbf{(b)} Same as panel \textbf{(a)}, with the position decomposed as the projection along six axes; the schematic of colored arrows in the inset matches the colored curves in the plot. 
    \textbf{(c)} Same as panel \textbf{(a)}, with the position decomposed as the distance to four points; see schematic in the inset.
    \textbf{(d)} Without any decomposition there is only one channel for the full input and we recover the standard Information Bottleneck.  
    The image degrades with increasing $\beta$, but the scheme of approximations grants far less insight about the relationship between position and color than the scheme of approximations evinced by the Distributed IB in panels \textbf{(a-c)}.
    }
    \label{fig:monalisa}
\end{figure*}

Peering into the black box allows us to connect the approximations with the circuitry: $X_3$ routes through the fewest gates to $Y$, making it particularly influential in determining the output.
The next coarsest approximation to exist for a significant range of $\beta$ adds information from $X_6$, $X_8$, $X_9$, and $X_{10}$.
These inputs all route through the same \texttt{XOR} gate late in the circuit; notably the Distributed IB does not compress them all identically even though \texttt{XOR} is commonly used as an example of a function where information about the output only arises from information about both inputs.
Instead, $X_{10}$ is informative about $Y$ without the rest of this subset, arising from the fact that the other input to this particular \texttt{XOR} will be \texttt{True} more often than not.
By consuming only input-output samples, the Distributed IB yields a rich signal about the precise means by which components determine the output.

In order to evaluate the quality of the approximation scheme found by the Distributed IB, we exhaustively measure the mutual information between all subsets of input components and the output $Y$ (Fig.~\ref{fig:circuit}c, black points).
To be concrete, there are ten subsets of a single input (one for each input gate), 45 possible pairs of inputs, and so on, with each subset sharing mutual information with $Y$ based on how the inputs are routed inside the black box.
The combinations of inputs that comprise the Distributed IB approximation scheme are the most informative subsets of their size (Fig.~\ref{fig:circuit}c, open markers).
The Distributed IB required only a single sweep---no exhaustive search through all subsets of inputs---to find a solution to the machine learning problem of feature subset selection: selecting the most informative subsets of features with regards to an output \cite{cai2018featureselectionreview}.
Importantly, there is also a full continuum of compressions of the input components between the discrete subsets, a rich signal not possible with classical feature subset selection~\cite{battitifeatureselection1994,peng2005featureselection}.

\subsection{Images: relation between position and color}
We seek new insight into the famous relationship between position and color in Leonardo da Vinci's painting of the Mona Lisa, by way of approximations found by the Distributed IB.
By specifying the input $X=\vec{r}$ as a vector in two-dimensional space, there is freedom of choice in its decomposition.
We begin in Fig.~\ref{fig:monalisa}a with a straightforward decomposition: the horizontal and vertical components of $\vec{r}$ as $X_1$ and $X_2$.
As horizontal and vertical information becomes more heavily compressed, approximations consist of coarsening blocks of color representing a shrinking number of distinct interactions between the two components.
The KL divergence traces reveal that vertical information explains more of the color than horizontal information, as the vertical component is maintained after the horizontal component has been compressed away.

Redundancy of information in the input components does not pose a significant challenge for the Distributed IB.
In Fig.~\ref{fig:monalisa}b, the position is decomposed as the projection onto six different axes. 
Information from all components is not necessary for perfect knowledge of the input, but some components will be more economical than others for conveying information about the output color $Y$.
The more descriptive components emerge in the traces of $D_\textnormal{KL}(U_i||R)$ as the bottleneck tightens, eventually relying on two axes to define a scheme of diamond-like approximations to the painting.
Similarly, when the position is decomposed as a triangulation (Fig.~\ref{fig:monalisa}c)---the Euclidean distance to a set of points---the Distributed IB finds an approximation scheme with a qualitatively different nature composed with gradually simplifying interactions between the triangulation components.

Finally, we asked how the approximations produced by the Distributed IB compared to those produced by the standard IB (Fig.~\ref{fig:monalisa}d).
The latter may be seen as a trivial decomposition of the input with the identity operator.  
Because the complete position can be encoded directly into its color before passing through the bottleneck, the scheme of approximations is uninformative because the degradation is only in the predicted color (see further discussion in Appendix B).
In comparing the approximation schemes of the Distributed IB and the standard IB, we find that only the former reveals insight about the relationship, through a prioritization of interactions between components of the input in determining the output. 

\begin{figure*}
    \centering
    \includegraphics[width=\textwidth]{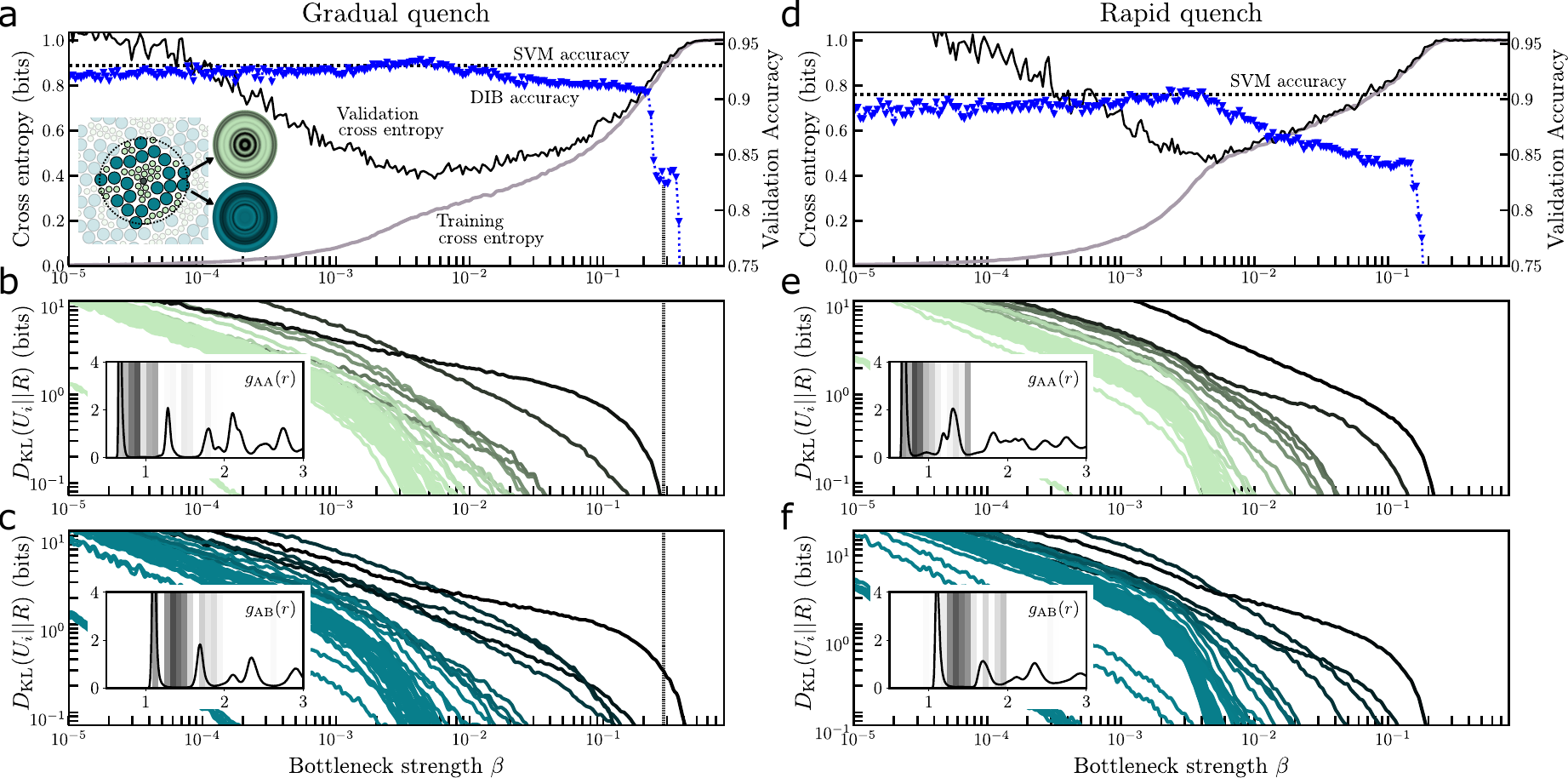}
    \caption{\textbf{Approximating the relationship between static structure and imminent rearrangement in a sheared glass.} \textbf{(a)} Distributed IB on a gradually quenched glass.
    \textit{Inset:} The local neighborhood around each particle is decomposed into 50 radial density values (``structure functions'') for each of the large and small particle types, for use in both the SVM and the Distributed IB.
    The classification task is to predict whether the center particle is the locus of an imminent rearrangement event.
    \textit{Main:} As in Figs.~\ref{fig:circuit}\&\ref{fig:monalisa}, the training error increases with the bottleneck strength $\beta$. 
    While $\beta$ is small, the training (black) and validation (gray) cross entropy errors (left vertical axis) display the hallmark behavior of overfitting. 
    Increasing $\beta$ for the Distributed IB acts as a regularizer and closes the gap between the two errors by restricting information about the structure functions. 
    The accuracy on the validation set (right vertical axis) of the Distributed IB (blue) is comparable with that of a support vector machine trained on the same data (dotted black) until a majority of the structure functions have been compressed away.
    As $\beta$ increases further, classification accuracy degrades until it drops precipitously, stopping briefly at an approximation scheme marked with the vertical gray bar.
    \textbf{(b)} The information maintained about each of the structure functions over the $\beta$ sweep from panel \textbf{(a)} for the smaller type A particles.
    \textbf{(c)} Same as panel \textbf{(b)} for the larger type B particles.
    The curves are colored by the order with which they fall below a threshold value of the $D_\textnormal{KL}(U_i||R)$.
    \textit{Insets:} The radial distribution function $g(r)$ measures the average density of radial shells for all particles in the system, with $g_\textnormal{XY}(r)$ the density of particles of type $\textnormal{Y}$ around a particle of type $\textnormal{X}$ at the origin.
    The radii are shaded according to the color scheme from the main panels of \textbf{(b,c)}, showing that the structure functions utilized for the coarser approximations by the Distributed IB are primarily those in the first troughs of $g_\textnormal{AA}(r)$ and $g_\textnormal{AB}(r)$. 
    The coarsest approximation, highlighted in panel \textbf{(a)} and classifying with nearly 85\% accuracy, utilizes information from only a single structure function involving the density of type B particles in the first trough of $g_\textnormal{AB}(r)$. 
    \textbf{(d-f)} Same as panels \textbf{(a-c)}, with a glass prepared via rapid quench.
    Beyond poorer classification accuracy for both the SVM and Distributed IB, the Distributed IB sweep over $\beta$ reveals how the connection between structure and rearrangement depends more strongly on information from many structure functions than in the gradual quench system.
    There is also no coarse approximation plateau as there was in panel \textbf{(a)}: with less information about the structure functions, all predictability of imminent rearrangement quickly degrades.
    Again, the insets of \textbf{(e-f)} show the most relevant structure functions lie in the troughs of $g_\textnormal{AA}(r)$ and $g_\textnormal{AB}(r)$ for the rapidly quenched glass.
}
    \label{fig:glass}
\end{figure*}

\subsection{Amorphous plasticity: relation between static configuration and future rearrangement}

Plastic deformation in disordered systems often occurs with intermittent rearrangement events~\cite{argon2013strain,murphy2019transforming,ridout2021avalanche}.
A longstanding question in the study of amorphous plasticity asks what markers in the static configuration of the system's elements predict future rearrangement dynamics~\cite{richard2020indicators,teich2021crystallinity}. 
To shed new light on this relationship between structure and dynamics, we train the Distributed IB on simulation data from Ref.~\cite{barbot2018simulations} of an athermal two-dimensional bidisperse Lennard-Jones glass under simple shear.
The particles that initiated sudden rearrangement events have been identified by the authors of Ref. \cite{richard2020indicators} as those contributing most to the critical mode at the onset of rearrangement (see Appendix A).
Our goal is to identify the rearrangement initiators from the rest of the particles in the system given only the static configuration of the local neighborhood.

We build upon an inventive machine learning approach that tackled this problem by first decomposing the local neighborhood into a fixed set of $\mathcal{O}(10^2)$ structure functions measuring the radial densities around a particle~\cite{behler2007structurefns}, and then training a support vector machine (SVM) to classify~\cite{cubuk2015PRL,schoenholz2016natphys,schoenholz2017pnas}.
The SVM approach achieves remarkable accuracy, and has since been extended to predict dynamics in a number of other amorphous systems~\cite{softnessGrainBoundaries,softnessFilms}.
Through knockout tests and the inspection of weights, the authors found that the predictive power primarily arose from only a few radial density values: those measuring the density of the closest shell of particles, which is located at the first peak of the radial distribution function $g(R)$.

We seek new insight through the use of the Distributed IB, by comparing the approximation schemes found for glasses prepared under different quench protocols (Fig.~\ref{fig:glass}).
To decompose the local neighborhood, we use 100 radial density values---50 for each of the two particle sizes (Fig.~\ref{fig:glass}a, inset).
We observe that the error in predicting $Y$---whether a particle is a locus of imminent rearrangement---climbs as the bottleneck strength $\beta$ increases (Fig.~\ref{fig:glass}a\&d).
The small $\beta$ regime exhibits the hallmark behavior of overfitting~\cite{lawrence2000overfitting}: the training error is nearly zero while the error on a held-out validation set is large.
Remarkably, as the bottleneck tightens, the validation error more closely matches the training error; presumably the information specific to samples of the training set is too costly for use in coarser approximations, suggesting that the bottlenecks serve as a regularization to inhibit overfitting.
Whereas common methods of protecting against overfitting in deep learning, such as dropout and L1/L2 regularization, encourage functional simplicity by limiting the number or magnitude of participatory weights \cite{ng2004regularization,wager2013dropout,zhang2021generalization}, the Distributed IB encourages simplicity by penalizing the information used in interactions between input components.

After the classification accuracy on the validation set peaks---at around the same value as the SVM---information about the structure functions starts to be discarded \emph{en masse} (Fig.~\ref{fig:glass}b,c,e,f).
The large majority of the 100 structure functions are removed from the approximations with the loss of only a few percent of classification accuracy on the validation set.
To classify with better than 80\% accuracy, information is needed about only a handful of structure functions that correspond to the troughs in the radial distribution functions, $g_\textnormal{AA}(r)$ and $g_\textnormal{AB}(r)$ (Fig.~\ref{fig:glass}b,c,e,f), in contrast to the finding of~\cite{schoenholz2016natphys} that the peaks were the most informative.
Rather than manually ablating or training on all possible subsets of structure functions~\cite{schoenholz2016natphys}, Distributed IB finds the relative information allocation between all structure functions, along with the corresponding predictability of $Y$ along the continuum.

By finding a series of approximations to the relationship between static structure and imminent rearrangement in these different glasses, we learn about fundamental differences between these complex systems.
Both the SVM and Distributed IB attain higher classification accuracy for the glasses that are gradually quenched (Fig.~\ref{fig:glass}a-c) compared to those that are rapidly quenched (Fig.~\ref{fig:glass}d-f).
The Distributed IB reveals a more significant deterioration of accuracy for the rapid quench when information about the structure functions decreases, indicating that the propensity for rearrangement depends more strongly on multiple signatures of the local structure than in the gradual quench.
For the gradual quench, there is an approximation that achieves nearly 85\% classification accuracy with information from only a single structure function, without an analogous approximation for the rapidly quenched glasses.
The approximation schemes for the two kinds of glasses reveals a simpler relationship between static structure and imminent rearrangement for the gradually quenched glass.

\section{Discussion}
The impressive ability of deep learning to find patterns in data has had limited value in science because interpretability is hard-earned and achieved on a case-by-case basis, if at all.
We have found that constraining the sum total of information incorporated from multiple components of an input in relation to an output serves to break down a relationship found by deep learning into comprehensible steps.
The information-theoretic erosion of a relationship into approximations brings interpretability to the black-box nature of deep learning and illuminates the explanatory structure of complex systems.

Statistical methods that reduce the input space of a relationship to important components have a long history.
Canonical correlation analysis (CCA) finds transformations of the input space that maximize linear correlation with the output~\cite{hotelling1936cca}, with extensions based on kernels~\cite{hardoon2004kcca} and deep learning~\cite{andrew2013deepCCA}.
Analysis of variance (ANOVA) methods decompose an input to parts that account for the most variation in the output~\cite{fisher1936anova,stone1994fanova,scheffe1999anova}, similarly with extensions to
deep learning~\cite{martens2020fANOVA}.
The Distributed IB uses deep learning to find compressed representations of all input components, and optimizes an objective based on the mutual information with the output.
Unique to the Distributed IB is the full approximation scheme: simulacra of the relationship that leverage decreasing amounts of information about the input and open a window to the nature of the relationship.

There is growing appreciation of the insight an information theory perspective can grant an analysis of physical systems.
The size of the lossless compression of a system state has been used to uncover meaningful order parameters~\cite{martiniani2019quantifying} and compute correlation lengths~\cite{martiniani2020correlation}.
Mutual information between partitions of a system allowed the automatic discovery and information-based prescription of relevant features~\cite{beny2018features,gokmen2021RSMI,koch2018natphys}.
The connection between the information bottleneck and the renormalization group~\cite{gordonrelevance2021,kline2021RGIB} suggests IB can uncover the most relevant information in a relationship.
By constraining the encoder in a standard IB framework to be a linear projection, the authors of Ref.~\cite{wang2019PIB} were able to glean influential parameters in biomolecular reactions.
While all of these works build upon the premise that tracking information in a system is a powerful means of understanding it, the distribution of bottlenecks is the critical step for directly measuring the importance of input components and finding meaningful approximations.

Inventive modifications to standard deep learning methods have brought an element of interpretability for the distillation of insight about systems in science.
SciNet~\cite{iten2020discovering} trains on specially formatted data taking the form of question-answer pairs about physical systems, in an autoencoder architecture modeled after human reasoning.
The artificial intelligence physicist~\cite{wu2019aiphysicist} is a sprawling framework connecting machine learning with strategies that have guided human physicists throughout history.
Another class of solutions finds economical descriptions of data out of a dictionary of possible functional forms: a process termed symbolic regression~\cite{koza1994genetic,schmidt2009distilling,udrescu2020aifeynman,wang2019symbolic}.
The Distributed IB operates in latent space and places no constraints on the architectures used to encode into or decode from the latent space.
The information factoring into the approximate relationships from the input components is constrained, though without knowledge of or constraints on the nature of the interactions between components.

\subsection{Methodological considerations}
When using the Distributed IB, it is important to consider several limitations.
The learned compression schemes for each input component, and the means by which the representations $\{U_i\}$ are integrated to predict $Y$, are found by deep learning and consequently inherit a lack of interpretability in exchange for greater functional complexity.
Second, there is freedom in the specific decomposition of $X$, bringing domain expertise and careful analysis center-stage in the application of the Distributed IB.
Finally, as in the Variational IB~\cite{alemiVIB2016}, the ability to integrate with deep learning by way of the variational objective (Eqn.~\ref{eqn:vib}) comes at the expense of any guarantees about the optimality of the representations in terms of Eqn.~\ref{eqn:DIB}.

\subsection{Conclusion}
By constraining the amount of information a deep learning architecture can utilize, we bestow upon it the ability to find a continuum of approximate relationships and in so doing, convey the nature of the relationship itself.
The fact that approximation plays a fundamental role in science underlies the significance of the Distributed IB to bring deep learning solidly into the scientists’ set of essential tools, and to accelerate the distillation of insight from data.

\section{Acknowledgements}
We gratefully acknowledge Dr. Sam Dillavou, Jenny Hamer, Dr. Erin G. Teich, Shubhankar Patankar, and Dr. Jason Z. Kim for helpful discussions and comments on the manuscript, and Dr. Sylvain Patinet for the amorphous plasticity data.

\section{Citation diversity statement}
Science is a human endeavour and consequently vulnerable to many forms of bias; the responsible scientist identifies and mitigates such bias wherever possible.
Meta-analyses of research in multiple fields have measured significant bias in how research works are cited, to the detriment of scholars in minority groups~\cite{chakravartty2018communicationsowhite,dion2018gendered,dworkin2020extent}.
We use this space to amplify studies, perspectives, and tools that we found influential during the execution of this research~\cite{zurn2020citation,dworkin2020citing,zhou2020gender,budrikis2020growing}.

\section{Appendix A: Implementation specifics}
All code was written in Tensorflow and will be released on Github at the time of publication.
All experiments used the Adam optimizer with a learning rate of $3\times10^{-4}$.

It has been shown that \textit{positionally encoding} low-dimensional features helps neural networks to learn high-frequency patterns \cite{tancik2020fourier}.
Because the decomposition of $X$ necessary for the Distributed IB creates multiple low-dimensional features, we found it helpful to positionally encode all continuous-valued features.
This procedure amounts to a Fourier mapping, taking each value $z$ to $[\textnormal{sin}(\omega_1 z), \textnormal{sin}(\omega_2 z), ...]$ where $\omega_k$ are the frequencies for the encoding.
We used $\omega_k=2^k\pi$, $k=\{1, 2, ..., k_\textnormal{max}\}$ along with the original feature $z$, such that every place where $z$ would be input instead received the vector $[z, \textnormal{sin}(\omega_1 z), \textnormal{sin}(\omega_2 z), ...]$.

While a separate encoder for each input component could be used for each experiment in this paper, we instead used one encoder for all components with a one-hot vector concatenated to indicate the components' identities.
With this approach, we found no negative effect on performance, and obtained faster runtimes and better reproducibility.
This solution is most sensible when the components are similar in nature, as in this work.

\subsection{Boolean circuitry implementation}
As each input may take only one of two values (0 or 1), the encoders were trainable constants $(\vec{\mu}_i,\textnormal{log}\ \vec{\sigma}_i^2)$ that were used to encode $p(u_i|x_i)= \mathcal{N} ((2x_i - 1)\times\vec{\mu}_i, \vec{\sigma}_i^2)$.
The decoder was a multilayer perceptron (MLP) consisting of three fully connected layers with 128 \texttt{tanh} units each.
We increased the value of $\beta$ logarithmically from $10^{-4}$ to $0.3$ in $2\times10^5$ steps after $10^4$ pre-training steps at the smallest $\beta$, with a batch size of 512 input-output pairs sampled randomly from the entire 1024-element truth table.

\subsection{Mona Lisa implementation}
The painting was resized to $600\times400$ RGB pixels.
The pixel grid was scaled to a $2\times\frac{4}{3}$ rectangle (maintaining the aspect ratio) centered at the origin to obtain the raw position $\vec{r}$ for each pixel.
The two-dimensional $\vec{r}$ was converted for the specific decompositions used in Fig.~\ref{fig:monalisa} (e.g., projected onto each of the six axes in Fig.~\ref{fig:monalisa}b).
These components were then positionally encoded with $k=\{1,2,...,9\}$.
A one-hot vector the size of the number of components $m$ was appended to each input, and all $m$ inputs were fed through an MLP of 5 layers of 512 \texttt{ReLU} units each.
The embedding dimension for each input component was 32.
After the distributed embeddings were obtained, they were concatenated for input to a combined encoder of 3 layers of 128 \texttt{ReLU} units each to embed to 64 dimensions.

The RGB color of a pixel was encoded to 64 dimensions with an MLP consisting of 3 layers of 128 \texttt{ReLU} units each.
The InfoNCE loss in the shared 64-dimensional embedding space used negative Euclidean (L2) distance as its similarity measure $s(u,v)$, and a temperature of 1.
With a batch size of 2048, training consisted of a logarithmic sweep over $\beta$ for $10^5$ steps from $10^{-6}$ to 3 after $10^4$ pre-training steps at the smallest $\beta$. 

To reconstruct an encoded image, 1024 colors were selected at random from the original image and embedded for use as a codebook for nearest neighbor retrieval.
For the full grid of pixels, each position was encoded and matched with its nearest color from the codebook in the shared embedding space, equivalent to finding the color (out of the 1024 sample colors) with maximum probability in the predicted distribution.

\subsection{Glassy rearrangement implementation}
The simulated glass data comes from Ref.~\cite{richard2020indicators}: 10,000 particles in a two-dimensional cell with Lees-Edwards boundary conditions interact via a Lennard-Jones potential, slightly modified to be twice differentiable~\cite{barbot2018simulations}.
Simple shear was applied with energy minimization after each step of applied strain.
The critical mode was identified as the eigenvector---existing in the $2N$-dimensional configuration space of all the particles' positions---of the Hessian whose eigenvalue crossed zero at the onset of global shear stress decrease.
The particle identified as the locus of the rearrangement event had the largest contribution to the critical mode~\cite{richard2020indicators}.

We used data from the gradual quench (``GQ'') and rapid quench (high temperature liquid, ``HTL'') protocols. 
Following Ref.~\cite{schoenholz2016natphys}, we considered only neighborhoods with type A particles (the smaller particles) at the center.
Under the premise that information about imminent rearrangement lies primarily in local deviations from the radial distribution function $g_\textnormal{XY}(r)$, the relationship is more interpretable if there is only one relevant $g_\textnormal{XY}(r)$ for each set of structure functions.
To be specific, by restricting the focus to neighborhoods with type A particles at the center, the important structure functions of type A particles need only be understood in light of $g_\textnormal{AA}(r)$, and similarly for the type B structure functions in light of $g_\textnormal{AB}(r)$.
If both particle types were considered, the important structure functions would derive from deviations in some weighted average of $g_\textnormal{AA}(r)$ and $g_\textnormal{BA}(r)$, and again for $g_\textnormal{AB}(r)$ and $g_\textnormal{BB}(r)$.
For each of 6,000 rearrangement events with a type A particle as the locus, we selected at random another type A particle from the same system state to serve as a negative example.
Of this set of 12,000 examples, 90\% were used for training and the remaining 10\% were used as the validation set.

The local neighborhood of each example was decomposed using 50 radial density structure functions for each particle type, evenly spaced over the interval $r=[0.25, 4]$.
Specifically, for particle $i$ at the center and the set of neighboring particles $X$ of type A,
\begin{equation}
    G_A(i;r,\delta)=\sum_{j\in X}\textnormal{exp}(-\frac{(R_{ij}-r)^2}{2\delta^2}),
\end{equation}
\noindent where $R_{ij}$ is the distance between particles $i$ and $j$.
The same expression was used to compute $G_B$, the structure functions for the type B particles in the local neighborhood.
We used $\delta$ equal to 50\% of each radius interval.

After computing the 100 values summarizing each local neighborhood, we normalized the training and validation sets with the mean and standard deviation of each structure function across the training set.
The normalization was performed to help the SVM approach, which performs best when the input features are all of similar scale~\cite{schoenholz2016natphys}.
The best validation results from a logarithmic scan over values for the $C$ parameter were used for the value of the SVM accuracy in Fig.~\ref{fig:glass}a,d.

For the Distributed IB, each of the 100 scalar values for the structure functions was positionally encoded with $k=\{1,2,3,4\}$ and concatenated with a 100-element one-hot vector, for input to an MLP consisting of 3 layers of 256 units with \texttt{ReLU} activation.
The embedding dimension of each $U_i$ was 64.
Then the 100 embeddings were concatenated for input to the decoder, which was another MLP consisting of 3 layers of 256 units with \texttt{ReLU} activation.
The output was a single logit to classify whether the particle at the center is the locus of imminent rearrangement.
We increased $\beta$ in equally spaced logarithmic steps from $10^{-6}$ to $2$ over 30,000 steps after 2,000 steps of pre-training at the smallest $\beta$.
The batch size was 512.

\section{Appendix B: The approximations of the Distributed IB and standard IB on binary distributions in 2D}
\begin{figure*}
    \centering
    \includegraphics[width=\textwidth]{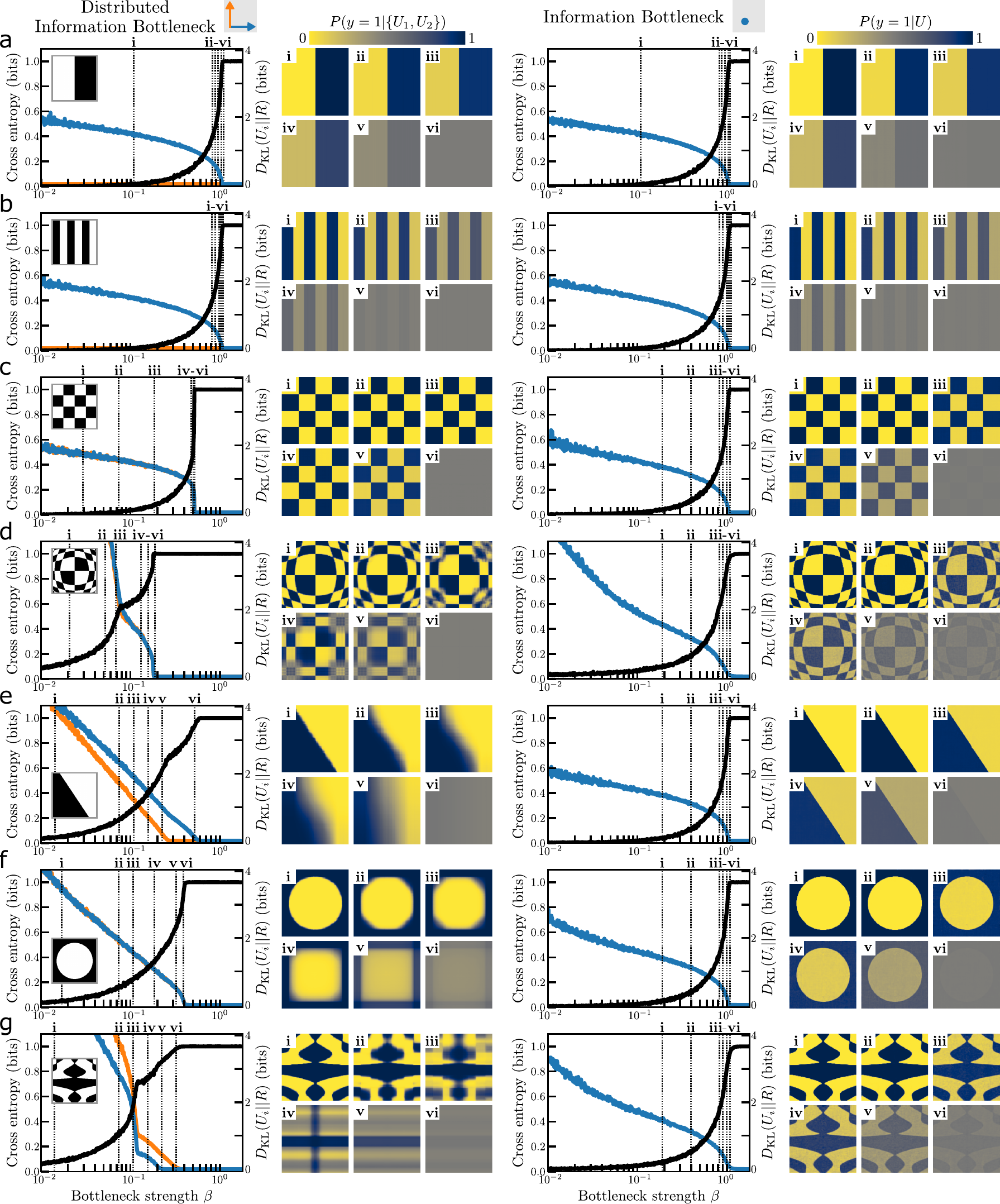}
    \caption{\textbf{Distributed, standard Information Bottleneck comparison on binary images with matching entropy.} 
    \textbf{(a)-(g)} \textit{Inset}: The binary images are distributions with continuous input $X$ over the unit square in $\mathbb{R}^2$, and output $Y \in \mathbb{B}$.
    The entropy $H(Y)=1$ bit for all images.
    The predicted output is a distribution $p(Y|U)$ with $U$ the representation for a position $X$, which can be displayed with a single value, $p(Y=1|U)$.
    For each row, we train the Distributed IB with horizontal and vertical components of $X$ as $\{X_1, X_2\}$ (left) and standard IB (right).
    We show the prediction error on $Y$ (cross entropy, black) and the relevant KL divergences (right, blue and orange) over the $\beta$ sweep, as well as noteworthy approximations indicated by the lowercase Roman numerals.}
    \label{fig:binary_supp}
\end{figure*}

Here we highlight the critical difference between the Distributed IB and the standard IB in terms of the former's ability to analyze relationships by way of a meaningful approximation scheme.
We create a series of binary images to analyze with both methods (Fig.~\ref{fig:binary_supp} Insets).
The support of the input $X$ is the unit square around the origin, and the output $Y \in \mathbb{B}$ has the same entropy for all images: $H(Y)=1$ bit.
The images represent a variety of complexity so we expect an informative analysis of the relationships to be able to differentiate between them.
We show that the Distributed IB yields a rich signal about the nature of these images, whereas the standard IB cannot meaningfully distinguish between the set of images.

For a fair comparison between the bottleneck methods, we trained neural networks with matching architectures and training parameters on the variational forms of both the Distributed IB and the standard IB, using the horizontal and vertical axes as the input decomposition for the former.
The encoder(s) and decoder were 3 fully connected layers of 128 \texttt{ReLU} units, with an embedding dimensionality of 8 for $U$.
In the case of the Distributed IB, the dimensionality of each $U_i$ was 4 so that the combined dimensionality (for the input to the decoder) matched that of the IB.
We swept $\beta$ logarithmically from $3\times10^{-4}$ to $3$ over 50,000 steps, with no pre-training.
The learning rate was $3\times10^{-4}$, with a batch size of 1024 input points sampled uniformly over the unit square for every mini-batch.
The horizontal and vertical components of the input were positionally encoded with $k=\{1,2,...,9\}$. 

In Fig.~\ref{fig:binary_supp}a-g, the $\beta$ sweeps for the Distributed IB reveal a rich variety of $\beta$ dependence of the prediction error and KL divergences.
In stark contrast, those for the IB are nearly identical for all images, with the information content of the representation $U$ dropping at $\beta=1$ across the board.
Because the standard IB processes the full input $X$, the representation $U$ is optimal if it sorts the input space into two clusters corresponding to $Y=0$ and $Y=1$, for all $\beta<1$; $\beta=1$ corresponds to the point where $I(X;U)=I(U;Y)=H(Y)$.
This fact may also be seen in the approximations for the IB, which are the trivial degradation of perfect reproductions. 
Thus the decomposition of $X$, and then the regulation of information flow about the components, is critical to the process of acquiring insight about the relationship.

There is much to say about what the Distributed IB finds for the different binary images.
For the images of Fig.~\ref{fig:binary_supp}a,b, perfect knowledge of $Y$ requires information about only the horizontal component of $X$: the KL term for the vertical component is always zero and everything else about the $\beta$ sweep matches that for the standard IB.
There is no sense of a frequency dependence: the Distributed IB does not distinguish between Fig.~\ref{fig:binary_supp}a and Fig.~\ref{fig:binary_supp}b in the same way that the IB does not distinguish between any of the images.
Instead, the Distributed IB measures the information allocated to different components of $X$ for predicting $Y$, without monitoring how that information is processed.

The remaining images contain interactions between the components of $X$, and the Distributed IB yields informative signals about each.
The checkerboard of Fig.~\ref{fig:binary_supp}c requires a single bit of information about each component of $X$ to determine $Y$.
This property of the image is manifest in where the approximation scheme ends: instead of ending at $\beta=1$ when 1 bit was needed about one component in Fig.~\ref{fig:binary_supp}a,b, the scheme ends at $\beta=0.5$ identically for both components.
This value of $\beta$ is where the terms of the Distributed IB loss are equal, $0.5(I(X_1;U_1)+I(X_2;U_2))=I(\{U_1,U_2\};Y)=H(Y)=1$ bit (assuming that the variational bounds are a good approximation for the mutual information values).
More information about the components of $X$ is required for the warped checkerboard of Fig.~\ref{fig:binary_supp}d, as indicated by the smaller $\beta$ where the approximation scheme ends.
There are two natural levels of approximation for this image, as shown in the error, KL divergences in the $\beta$ sweep, and the notable approximation images.
In the first robust approximation, the rounded boundaries to the checkerboard are maintained but at a high information cost, and after $\beta\approx 6\times10^{-2}$, a crude rectilinear approximation is adopted.

The slanted partitioning of input space for the image of Fig.~\ref{fig:binary_supp}e, while intuitively simple, is not able to be parsimoniously expressed in the particular decomposition of $X$ into horizontal and vertical components.
The same holds for the circle of Fig.~\ref{fig:binary_supp}f.
Information about both components of $X$ is needed to specify the value of $Y$, and optimization of the Distributed IB gradually coarsens the approximation toward a rectilinear scheme as information costs more (through increasing $\beta$).
For the slant of Fig.~\ref{fig:binary_supp}e, the KL divergence terms and the notable approximations show how vertical information is eventually compressed away, and all that remains is a horizontal gradient in the prediction of $Y$.
The most visually complex image (Fig.~\ref{fig:binary_supp}g) is approximated with a sequence of qualitatively distinct steps, culminating in one utilizing only the vertical component of the input $X$.
No such information is revealed by the standard IB sweeps of the same images: only by regulating information flow about components of $X$ do we obtain a meaningful analysis of the relationship between $X$ and $Y$.

% \bibliography{references}
%merlin.mbs apsrev4-1.bst 2010-07-25 4.21a (PWD, AO, DPC) hacked
%Control: key (0)
%Control: author (8) initials jnrlst
%Control: editor formatted (1) identically to author
%Control: production of article title (-1) disabled
%Control: page (0) single
%Control: year (1) truncated
%Control: production of eprint (0) enabled
%

\end{document}